%% file: Tams.tex
\documentclass[11pt]{TinLatex} 
\usepackage{graphicx}
\usepackage{epstopdf}

\usepackage{amsmath,amsxtra,amssymb,latexsym, amscd}

\usepackage[mathscr]{eucal}

\usepackage{hyperref}
\hypersetup{
 colorlinks=true,
 linkcolor=blue,
}

\usepackage{xspace}
\usepackage{algorithmic}
\usepackage{graphicx}
\usepackage{textcomp}
\usepackage{tabularx,ragged2e}
\usepackage{indentfirst}
\usepackage{subcaption}
\usepackage{colortbl}
\usepackage{diagbox}
\usepackage[table,xcdraw]{xcolor}
\usepackage{amsfonts}
\usepackage{booktabs}
\usepackage{siunitx}
\usepackage{array}
\usepackage{amsmath}
\usepackage{float}
\usepackage{booktabs, caption, makecell}
\usepackage[symbol]{footmisc}

\usepackage{threeparttable}
\usepackage[linesnumbered,ruled]{algorithm2e}

\font\it=cmti12 at 11pt
\font\bf=cmbx12 at 11pt

\advance\hoffset0.4cm
\font\bfh=cmssbx10 scaled 1400
\font\bfl=cmssbx10 scaled \magstep1

\font\chu=cmr8

\font\cn=cmr10

\def\vb{\vskip0.17cm}

\def\ed{\end{document}}
\def\vuong{\raise-0.16cm\hbox{$^\blacksquare$}}

\makeatletter
\DeclareRobustCommand\onedot{\futurelet\@let@token\@onedot}
\def\@onedot{\ifx\@let@token.\else.\null\fi\xspace}

\makeatother

\usepackage{cite}
\usepackage{multirow}
\usepackage[numbers]{natbib}
\usepackage{tablefootnote}

\begin{document}
\makeatletter	   
\setcounter{page}{1}
	

\def\footnoterule{\kern-3\p@
	\hrule \@width 1.32in \kern 2.6\p@} 
\renewcommand{\footnotemargin}{0em}
\newcommand{\fn}[1]{\footnotetext{\hspace{-6mm}#1}}
\setlength{\skip\footins}{8mm}

\makeatother   


\title{Integrating Image Features with Convolutional Sequence-to-sequence Network for Multilingual Visual Question Answering}
\author{
	{\cn TRIET M. THAI, SON T. LUU$^*$}
	\vskip.5cm
	{
	\it University of Information Technology, Ho Chi Minh City, Viet Nam\\
	\it Vietnam National University, Ho Chi Minh City, Vietnam
\fn{*Corresponding author.}
\fn{\hspace{1.7mm}{\it E-mail addresses}: \href{mailto:19522397@gm.uit.edu.vn}{19522397@gm.uit.edu.vn} (T.M.Thai); 
	\href{mailto:SONLT@uit.edu.vn}{sonlt@uit.edu.vn} (S.T. Luu).
	}
}}

\maketitle
\renewcommand\refname{\normalsize \centerline{ REFERENCES}}
\pagestyle{plain}
\pagestyle{myheadings}
\markboth{\footnotesize \chu \uppercase{ TRIET M. THAI}, SON T. LUU}
{\footnotesize  \chu \uppercase{Integrating Image Features with Convolutional Sequence-to-sequence Network}}

\begin{abstract} Visual Question Answering (VQA) is a task that requires computers to give correct answers for the input questions based on the images. This task can be solved by humans with ease but is a challenge for computers. The VLSP2022-EVJVQA shared task carries the Visual Question Answering task in the multilingual domain on a newly released dataset: UIT-EVJVQA, in which the questions and answers are written in three different languages: English, Vietnamese and Japanese. We approached the challenge as a sequence-to-sequence learning task, in which we integrated hints from pre-trained state-of-the-art VQA models and image features with Convolutional Sequence-to-Sequence network to generate the desired answers. Our results obtained up to 0.3442 by F1 score on the public test set, 0.4210 on the private test set, and placed $3^{rd}$ in the competition.
\vb

\end{abstract}

\input{sections/1-introduction.tex}
\input{sections/2-relatedworks.tex}
\input{sections/3-dataset.tex}
\input{sections/4-method.tex}
\input{sections/5-experiments.tex}

\input{sections/6-conclusion.tex}

\section*{ACKNOWLEDGMENT}
We would like to thank and give special respect to VLSP organizers for providing the valuable dataset for this challenge. 



	{\small
		\bibliographystyle{IEEEtranS} 
		\bibliography{refs.bib} 
	}



\end{document}

%% file: sections/1-introduction.tex
\section{INTRODUCTION}
\label{intro}
Visual Question Answering (VQA) is a trending research topic in artificial intelligence that combines natural language processing and computer fields. This task enables computers to extract meaningful information from images and answer the question by natural language text. The VQA task has various practical applications such as chat-bot systems, intelligent assistance, and recommendation system. 

\begin{figure}[ht!]
\centering
  \includegraphics[clip,width=0.8\textwidth]{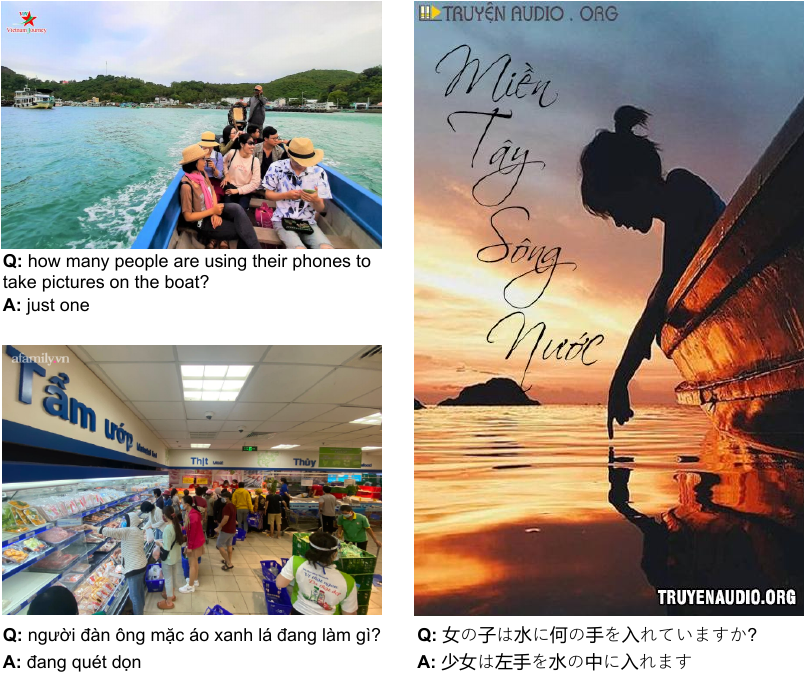}%
  \label{ex31}
\caption{Multilingual samples from UIT-EVJVQA dataset. From top to bottom, left to right: English (en), Vietnamese (vi) and Japanese (ja). The dataset contains a wide variety of questions; in some cases, the image contains noises that makes it difficult for a computer to distinguish the indicated object or action, for instance, ``phones" in the English example or the action of ``the man in green shirt" in the Vietnamese case. Besides, the Japanese example provides a tough scenario, "which hand is the girl putting onto the water?" in English, that even humans find it challenging to deliver the proper response.}
\label{example}
\end{figure}

The VQA can be categorized as a question-answering (QA) task. In the QA task, cross-lingual language QA has been a hot trend in recent years with the appearance of BERT \cite{devlin-etal-2019-bert} (trained on more than 100 languages), as well as plenty of multilingual datasets \cite{10.1145/3560260}. The VLSP-EVJVQA challenge \cite{vlsp2022} takes the VQA as a multilingual QA task, containing three different languages: Vietnamese, Japanese, and English. The challenge brings the first large-scale multilingual VQA dataset, UIT-EVJVQA, with approximately 5,000 images and more than 30,000 question-answer pairs. The task takes an image and a question with text form as input, and the computer must return the correct answer by text as output. The questions are written in Vietnamese, Japanese, or English, and the answers must follow the language used in the questions. To create the correct answer, the computer must understand the question content and extract the information from the corresponding image. Fig.\ref{example} illustrates several examples from the dataset provided by the organizer. According to the answer types, the VLSP-EVJVQA is a Free Form Answer QA task \cite{dzendzik-etal-2021-english}.

To solve the VLSP-EVJVQA, we propose our solution, which combines the Sequence-to-Sequence (Seq2Seq) learning with image features extraction task to generate the correct answers. We use the  ViLT \cite{pmlr-v139-kim21k} and OFA \cite{wang2022ofa} for hints extraction from the image, 
Vision Transformer \cite{dosovitskiy2020vit} for image features extraction 
and the Convolutional network for Seq2Seq learning \cite{10.5555/3305381.3305510} to generate the answer. Our results achieve the $3^{rd}$ place in the VLSP-EVJVQA. Hence, we describe our works in this paper. The paper is structured as follows. Section \ref{intro} introduces the task. Section \ref{relateds} takes a brief survey about previous works for the VQA task. Section \ref{dataset} overviews the VLSP-EVJVQA dataset. Section \ref{method} describes our proposed solution. Section \ref{results}devotes to experiments and performance analysis. Finally, Section \ref{conclusion} concludes our works and presents future studies.

%% file: sections/2-relatedworks.tex
\section{RELATED WORKS} 
\label{relateds}
\subsection{Existing datasets and methods for visual question answering}
In computer vision, the research purpose for VQA is to make computers understand the semantic context of images. The Microsoft COCO dataset \cite{lin2014microsoft} is one of the large-scale datasets that impact many studies in computer vision tasks, including object detection, image classification, image captioning, and visual question answering. Several VQA datasets are built on the MS-COCO in different languages, such as the VQA \cite{Antol_2015_ICCV}, VQAv2 \cite{Goyal_2017_CVPR} in English, FM-IQA \cite{10.5555/2969442.2969496} for Chinese, the Japanese VQA \cite{shimizu-etal-2018-visual} for Japanese, and the ViVQA \cite{tran-etal-2021-vivqa} for Vietnamese. There are also other two benchmark datasets for training and fine-tuning VQA methods, including 
Visual Genome (VG-QA) \cite{10.1007/s11263-016-0981-7} and GQA \cite{hudson2018gqa}. VG-QA is a VQA  dataset that contains real-world photographs. It is designed and constructed to emphasize the interactions and relationship between natural questions and particular regions on the images. The creation of VG-QA lays the groundwork for building GQA, another large VQA collection that make use of Visual Genome scene graph structures to feature compositional question answering and real world reasoning. Besides, in the natural language processing field, the SQuAD dataset \cite{rajpurkar-etal-2016-squad} has boosted many studies in question-answering and natural language understanding. Based on SQuAD, many corpora are created in different languages like DuReader \cite{he-etal-2018-dureader} for Chinese, JaQuAD \cite{so2022jaquad} for Japanese, KorQuAD \cite{lim2019korquad1} for Korean, and ViQuAD \cite{nguyen-etal-2020-vietnamese,van2022vlsp} for Vietnamese.

Apart from creating high-quality datasets, the architecture also plays a vital role in constructing intelligence systems. Taking advantage of natural language processing, we have several robust models for sequence-to-sequence learning tasks such as Long-short Term Memory \cite{8296600}, Convolutional Neural Networks for Sequence-to-sequence \cite{10.5555/3305381.3305510}, Transformer \cite{NIPS2017_3f5ee243} and BERTology \cite{rogers-etal-2020-primer}. In computer vision, state-of-the-art models for extracting useful information from images includes YOLO \cite{redmon2016you}, VGG \cite{simonyan2014very}, Vision Transformer (ViT) \cite{dosovitskiy2020vit}. With the increasing diversity of data and the need to solve multi-modal tasks that involve both visual and textual features, recent research trends focus on developing models that combine both the vision and language modalities such as: Vision-and-Language Transformer (ViLT) \cite{pmlr-v139-kim21k}, or OFA \cite{wang2022ofa}.

\subsection{Vision-language models}
\subsubsection{Vision-and-Language Transformer (ViLT)}

Introduced at ICML 2021, Vision-and-Language Transformer or ViLT \cite{pmlr-v139-kim21k} is one of the first and considerably, the simplest architectures that unifies visual and textual modalities. The model takes advantage of transformer module to extract and process visual features without using any region features or convolutional visual embedding components, making it inherently efficient in terms of runtime and parameters. The architecture of the ViLT model is originally set up to approach the VQA problem in the direction of a classification task. Thus, the output of the model contains various keyword answers with respect to probabilities.

\subsubsection{The "One For All" architecture that unifies modalities (OFA)}

OFA \cite{wang2022ofa} is a unified sequence-to-sequence pre-trained model that can generate natural answers for visual question answering task. The architecture uses Transformer \cite{NIPS2017_3f5ee243} as the backbone with the Encoder-Decoder framework. The model is pre-trained on the publicly available datasets of 20M image-text pairs and achieves state-of-the-art performances in a series of vision and language downstream tasks, including image captioning, visual question answering, visual entailment, and referring expression comprehension, making it a promising component in our approach toward the VQA challenge.

\subsection{Convolutional Sequence-to-sequence Network}

Sequence-to-sequence learning (Seq2Seq) is a process of training models to map sequences from one domain to sequences in another domain. Some common Seq2Seq applications include machine translation, text summarization and free-form question answering, in which the system can generate a natural language answer given a natural language question. 
A trivial case of Seq2Seq, where the input and output sequences are in the same length, can be solved using a single Long Short Term Memory (LSTM) or Gated Recurrent Unit (GRU) layer. In a canonical Seq2Seq problem, however, the input and output sequences are of different lengths, and the entire input sequence is required to begin predicting the target. This requires a more advanced setup, in which RNN-based encoder-decoder architectures commonly used to address the problem. The typical and generic architecture of these type of models include two components: an encoder that processes input sequence $X = (x_1, x_2,..., x_m)$ and return state representation $z =  (z_1, z_2,..., z_m)$, also known as context vector, and a decoder that decodes the context vector and outputs the target sequence $y = (y_1, y_2,..., y_n)$ by generating it left-to-right consecutively, one word at a time.

Different from other Seq2Seq models such as bi-directional recurrent neural networks (Bi-RNN) with soft-attention mechanism \citep{bahdanau2016neural, luong-etal-2015-effective} or the mighty Transformer \citep{NIPS2017_3f5ee243} with self-attention, the convolutional sequence-to-sequence network (ConvS2S) has the Encoder - Decoder architecture based entirely on convolutional neural networks and originally set up for machine translation task. The model employs many convolutional layers, which are commonly used in image processing, to enable parallelization over each element in a sequence during training and thus better utilize GPU hardware and optimization compared to recurrent networks. ConvS2S applies a special activation function called the Gate Linear Unit (GLU) \cite{DBLP:journals/corr/DauphinFAG16} as non-linearity based gating mechanism over the output of the convolution layer, which has been shown to perform better in the context of language modeling. Multi-step attention is also the key component of the architecture that allows the model to make multiple glimpses across the sequence to produce better output.

%% file: sections/3-dataset.tex
\section{THE DATASET}
\label{dataset}

The dataset released for the VLSP-EVJVQA challenge, UIT-EVJVQA \cite{vlsp2022}, is the first multilingual Visual Question Answering dataset with three languages: English (en), Vietnamese (vi), and Japanese (ja). It comprises over 33,000 question-answer pairs manually annotated on approximately 5,000 images taken in Vietnam, with the answer created from the input question and the corresponding image. Besides various types of questions, the answers are constructed in a free-form structure, making it a challenge for VQA systems. To perform effectively and achieve good results on UIT-EVJVQA, the typical VQA systems must identify and predict correct answers in free-form format for multilingual questions, due to the dataset characteristics.

\begin{table}[ht!]
    \centering
    \renewcommand{\arraystretch}{1}
    \resizebox{\textwidth}{!}{%
    \begin{tabular}{lrrrcrrr}
        \toprule
        & \multicolumn{3}{c}{Training set} && \multicolumn{3}{c}{Public test set} \\
        \cmidrule{2-4} \cmidrule{6-8}    
        & \textbf{English} & \textbf{Vietnamese} & \textbf{Japanese} &&  \textbf{English} & \textbf{Vietnamese} & \textbf{Japanese} \\
        \midrule
        Number of samples &7,193 &8,320 &8,261 &&1,686 &1,678 &1,651 \\
        \textbf{Questions} & \multicolumn{3}{c}{} & \multicolumn{3}{c}{}\\
        Vocabulary size &2,089 &1,860 &3,035 &&1,080 &919 &1,226 \\
        Average Length &8.52 &8.70 &13.03 &&8.76 &8.87 &13.27 \\
        Max Length &24 &21 &45 &&26 &22 &33 \\
        Min Length &3 &3 &4 &&3 &4 &4 \\
        \textbf{Answers} & \multicolumn{3}{c}{} & \multicolumn{3}{c}{}\\
        Vocabulary size &2,307 &2,067 &3,534 &&1,029 &877 &1,176 \\
        Average Length &5.09 &6.04 &7.27 &&3.89 &4.54 &5.05 \\
        Max Length &23 &23 &30 &&19 &18 &21 \\
        Min Length &1 &1 &1 &&1 &1 &1 \\
        \cmidrule{1-8} 
    \end{tabular}}
    \caption{Statistical information about the UIT-EVJVQA dataset}
    \label{statistic_dataset}
\end{table}

The training set and public test set have total samples of 23,774 and 5,015, respectively. Table \ref{statistic_dataset} describes the statistical information about the UIT-EVJVQA dataset on the training and public test sets. The length of sentences is computed at the word level. We use Underthesea\footnote{\url{https://github.com/undertheseanlp/underthesea}}
and Trankit \cite{nguyen2021trankit} library for word segmentation. 
Generally, the distribution of the training and public test sets is quite similar. English has fewer training samples compared with Vietnamese and Japanese, which may affect the question answering performance on this language. The length of questions is longer than the length of answers in three languages. The questions in Japanese are significantly longer than those in the two remaining languages. For the answers, those in Japanese are still also longer than those in English and Vietnamese. However, the difference in the length is not as much as the questions. Particularly, the shortest answers in three languages have only one word. In contrast, the questions and answers in Vietnamese have fewer words than those in English and Japanese.

%% file: sections/4-method.tex
\section{THE PROPOSED METHOD}
\label{method}
Figure \ref{fig_method} depicts an overview of the proposed approach in this study. In general, we transform the VQA problem into a sequence-to-sequence learning task, in which we take advantage of State-of-the-art (SOTA) vision-language models to offer richer information about the question-image dependencies in the input sequence.
 The method consists of two main phases that are carried out sequentially. At the first stage, numerous hints are extracted from question-image pairs using pre-trained vision-language models. The extracted hints are then concatenated with the question and visual features to form a sequence representation as input to the proposed Seq2Seq model to generate the corresponding answers in free-form natural language. Our source code and demo for the proposed methodology are available at this link: \url{https://huggingface.co/spaces/daeron/CONVS2S-EVJVQA-DEMO}

\begin{figure}[ht]
\centering
\includegraphics[width=\textwidth]{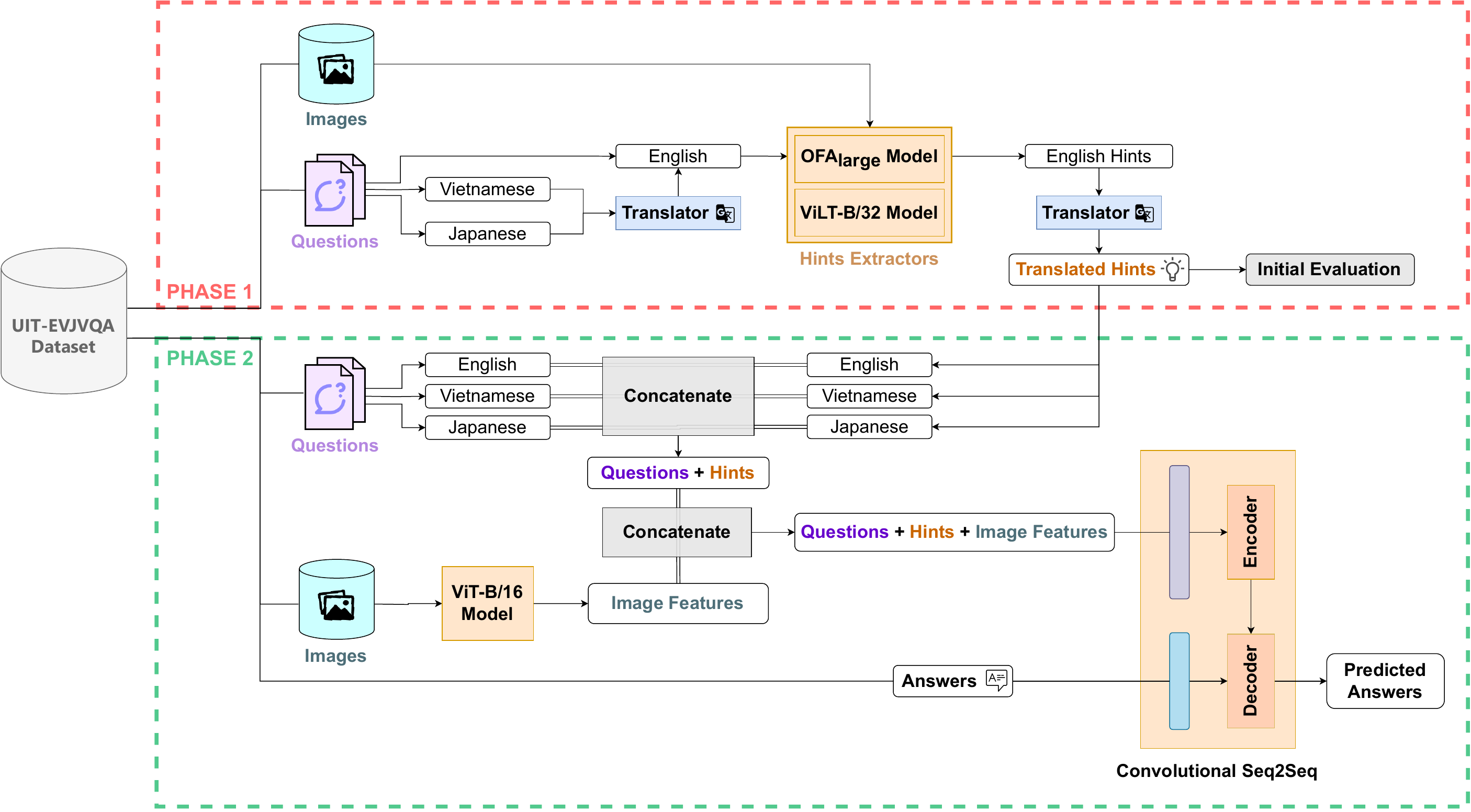}
\caption{An overview of the proposed method for visual question answering on UIT-EVJVQA dataset}
\label{fig_method}
\end{figure}

\subsection{Hints extraction with pre-trained vision-language models}

This phase concentrates on implementing SOTA vision-language models, including OFA \citep{wang2022ofa} and ViLT \citep{pmlr-v139-kim21k} to predict the possible answers given a question and its corresponding image. Due to the diverse nature of the questions and the multilingual aspect of UIT-EVJVQA, these models are only set up to provide answers directly through zero-shot prediction, with no training or fine-tuning step on the dataset. These SOTA models, which were pre-trained and fine-tuned on various datasets (VQAv2, VG-QA, and GQA), mainly support English but do not yet support Vietnamese or Japanese. To achieve desired results, we translate the Vietnamese and Japanese questions from UIT-EVJVQA into English using Google Translate API\footnote{\url{https://cloud.google.com/translate}} before feeding them into the models to get inferences. Once the output answers are generated, they are translated back into the original languages for evaluation and experiments in the second phase. For ViLT, we choose up to five candidate answers with the highest probability for further experiments. Using more hints is feasible, but it will put more pressure on computational resources as we approach creating long sequences based on hints probability in the next phase. We concatenate each output answer from ViLT along the sequence, respectively with decreasing relevance, to assess their quality on the new dataset. The inference performance of pre-trained ViLT and OFA models on the public test set are shown in Table \ref{vqa_example}.

Under our expectations, the OFA model with unified Seq2Seq structure outperforms ViLT with F1 0.1902, while ViLT achieves the best performance with F1 0.1317 using 2 keyword answers. The evaluation results are not quite good compared to the ground truth because of the special characteristic of the dataset with long answers and, since no training has been done, the predicted answers lack sufficient vocabulary. The predicted answer may match or not match the ground truth completely but gives a similar and proper response to the question. Regardless of accuracy, these simple keyword answers provide valuable insights about question-image interactions. Due to this, we consider these answers as hints or suggestions for each question-image pair and apply them to the training of the main model in the following phase.

\begin{table}
\centering
\begin{tabular}{lcc}
\toprule
\textbf{Model}                & \textbf{\# hints} & \textbf{F1}      \\ 
\midrule
\multirow{5}{*}{ViLT} & 1       & 0.1303  \\
                      & 2       & 0.1317  \\
                      & 3       & 0.1315  \\
                      & 4       & 0.1290  \\
                      & 5       & 0.1252  \\ 
\midrule
OFA                 & -      & 0.1902  \\
\bottomrule
\end{tabular}
\caption{Performance of SOTA vision-language models on public test set}
\label{vqa_example}
\end{table}

\subsection{Experiment with Convolutional Sequence-to-sequence Network}
The second phase of the approach concentrates on developing and training the main model for this challenge: the Convolutional Sequence-to-sequence Network (ConvS2S) \citep{10.5555/3305381.3305510} with different combinations of textual and image features for visual question answering task. ConvS2S has significant capabilities to accelerate training progress and reduce our computational resource limitations due to its efficiency in terms of GPU hardware optimization and parallel computation. This is why the architecture is preferred over other Seq2Seq models for the competition.

In this study, each convolutional layer of ConvS2S uses many filters with a width of 3. Each filter will slide across the sequence, from beginning to end, looking at all 3 consecutive elements at a time to learn to extract a different feature from the questions, hints, visual factors and answers. With these special settings, the model has a significant capacity to extract meaningful features from the input sequence and generate free-form content. Due to its proven performance in other Seq2Seq learning tasks, such as machine translation, we anticipate the model to perform well on question-image features combination and produce good results on the visual question-answering task.


\subsection{Textual and visual features combination}
In the early stage, a set of various useful hints is achieved using pre-trained ViLT and OFA. In order to train the proposed Seq2Seq model with the existing materials, the textual features, including questions and hints, and image features have to be combined in the form of sequence representations as input to the Seq2Seq model.

\begin{figure}[!htbp]
\centering
\includegraphics[width=\textwidth]{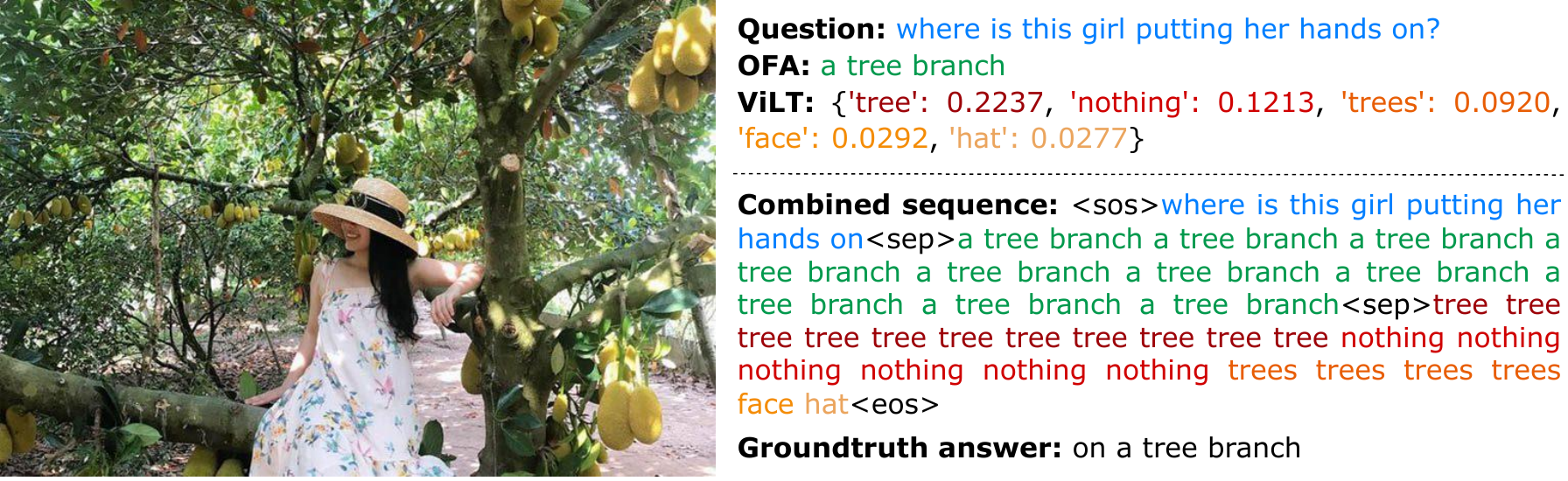}
\caption{An example of question and hints combination. The hint `tree' occurs 11 times in the sequence since the half of its probality is 11.19 (\%). }
\label{fig:combine}
\end{figure}

As shown in Table \ref{vqa_example}, adding more ViLT hints to the sequence tend to reduce the F1 score performance. These simple answers, on the other hand, may passively contribute to the overall understanding of the scenario of the corresponding images. Therefore, our approach focuses on using hint probability to generate sequences with repeated keywords while avoiding noise from outliers. This method allows hints that have a higher probability to appear more frequently in the sequence. For efficiency and reducing cost, the number of times a ViLT hint occurs in the sequence is the integral part of its half probability. For experiments involving the output of two models, the hint from the OFA model is set to appear 10 times in the sequence. The newly created sequence is concatenated with the question to form the final sequence for question and hint. We then remove special characters, lowercasing, and tokenize the text contents before passing them into the encoder. For English content, we tokenize the text simply by splitting them word-by-word. For Vietnamese and Japanese content, Underthesea toolkit and Trankit \citep{nguyen2021trankit} libraries are applied for word segmentation, respectively. Figure \ref{fig:combine} illustrates an example of a question and hints combination in our approach.

Besides the hints from question-image pairs, we also apply the Vision Transformer (ViT) \citep{dosovitskiy2020vit} to extract visual features from the image. The input image is passed into ViT model to obtain
a sequence of patches called the patch embeddings, which then pass through a Transformer encoder with multi-head attention to output the image features with the size of 196 x 768. Once the image features are achieved, we remove the vector at [CLS] token position and concatenate these visual features with text embeddings along the sequence dimension to have the final representative embedding matrices for questions, hints, and images.

%% file: sections/5-experiments.tex
\section{EXPERIMENTS AND ANALYSIS}
\label{results}
\subsection{Experiment Settings}
The ConvS2S model has 512 hidden units for both encoders and decoders. All embeddings, including the output produced by the decoder before the final linear layer, have a dimensionality of 768. This setup allows the encoders to concatenate with patch embeddings from ViT model. To avoid overfitting, dropout is applied on the embeddings, decoder output, and the input of the convolutional blocks with a retaining probability of 0.5. Teacher forcing with probability of 0.5 in the is also applied in the architecture to accelerate the training progress.

Many experiments are carried out in order to evaluate the proposed approach toward the VLSP-EVJVQA challenge. Typically, the training and evaluation of ConvS2S model is conducted using four types of input sequence: question only, question-image, question-hint and question-hint-image. First, we initialize the baseline result of ConvS2S with only question as input sequence and no image information. This scenario is similar to the Knowledge-based question answering (KBQA) task in that the generated answers are entirely based on the question-answer dependencies learned during the training phase. The second experiment involved image features combined with question as typical VQA approach. We then add visual hints to the input sequences used in the two prior experiments and investigate their effect on overall performance.

Because of the limitation in computational resources as well as the strict timeline of the competition, we only deploy the fine-tuned ViLT-B/32 with 200K pretraining steps and pre-trained OFA$_{\mathrm{large}}$ with 472M parameters for hints inference given the question and image. For feature extraction from image, we deploy pre-trained ViT-B/16 with base-sized version.
To have the comparative result, we set up the same hyperparameters for all experiments with ConvS2S. The model is trained in 30 epochs with batch size of 128 using Adam optimizer with a fixed learning rate of 2.50e-4. After each epoch, the performance loss on the train and development sets is calculated using the Cross-Entropy Loss function.

The proposed architecture and SOTA vision and language models are implemented in PyTorch and trained on the Kaggle platform with hardware specifications: Intel(R) Xeon(R) CPU @ 2.00GHz; GPU Tesla P100 16 GB with CUDA 11.4.

\subsection{Experimental Results}

\begin{table}[!htbp]
    \centering
    \resizebox{\columnwidth}{!}{%
    \setlength{\tabcolsep}{5pt}
    \renewcommand{\arraystretch}{1.2}
    \begin{tabular}{lcccccc}
    \toprule
        \textbf{Model} & \textbf{F1} & \textbf{BLEU-1} & \textbf{BLEU-2} & \textbf{BLEU-3} & \textbf{BLEU-4} & \textbf{BLEU}  \\ \midrule
        ConvS2S (Question only) & 0.3005 &0.2592	&0.2034	&0.1677	&0.1425& 0.1932  \\ \midrule
        ConvS2S + ViT & 0.3109 &0.2683	&0.2119	&0.1747	&0.1480 & 0.2007  \\ \midrule
        ConvS2S + ViLT & 0.3294 &0.2692	&0.2109	&0.1723	&0.1446& 0.1993  \\ 
        ConvS2S + OFA & 0.3331 &0.2858	&0.2269	&0.1876	&0.1598 & 0.2150  \\ 
        \textbf{ConvS2S + ViLT + OFA}
        & \textbf{0.3442} &0.2797	&0.2205	&0.1808	&0.1529 &0.2085  \\ 
        \midrule
        ConvS2S + ViT + ViLT & 0.3361 &0.2833	&0.2243	&0.1845	&0.1564 & 0.2122  \\ 
        ConvS2S + ViT + OFA & 0.3390 &0.2877	&0.2276	&0.1877	&0.1593 & \textbf{0.2156}  \\
        \textbf{ConvS2S + ViT + ViLT + OFA}
        & \textbf{0.3442} & 0.2747	&0.2148	&0.1747	& 0.1465& 0.2027 \\ \bottomrule
    \end{tabular}}
    \caption{Performance of ConvS2S with different combinations of pre-trained models on the public test set.}
    \label{tab:repub}
\end{table}


\begin{figure}[ht]
\centering
\includegraphics[width=\textwidth]{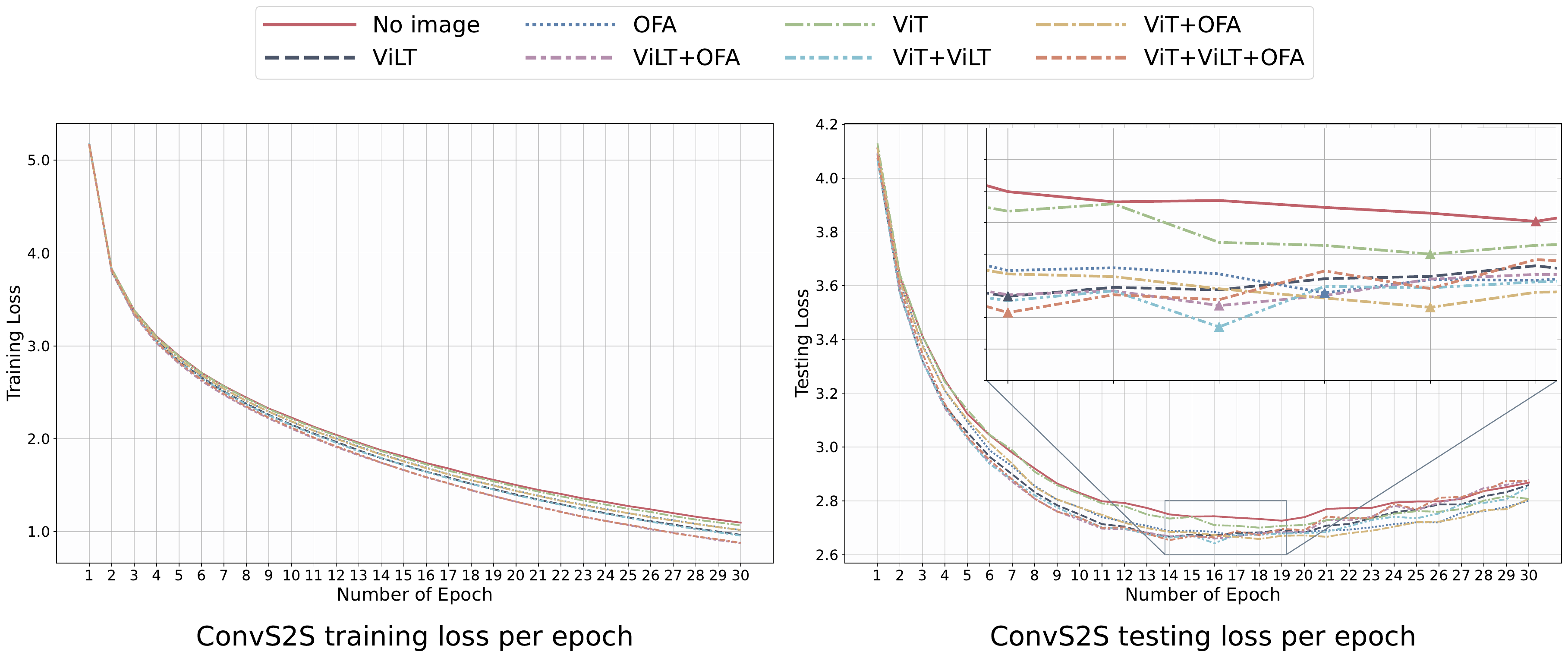}
\caption{Training loss and public testing loss comparison of ConvS2S model with different combinations of hint and image features.}
\label{loss}
\end{figure}

The two metrics: F1 and BLEU, are used in the challenge to evaluate the results. The BLEU score is the average of BLEU-1, BLEU-2, BLEU-3, and BLEU-4. F1 is used for ranking the final results. Table \ref{tab:repub} presents the performance of the proposed ConvS2S model with different combinations of pre-trained models on the UIT-EVJVQA public test set.

According to Table \ref{tab:repub}, the original ConvS2S model using only question obtained 0.3005 by F1 and 0.1932 by BLEU. Using question-image pairs, ConvS2S achieves a marginally better performance on both metric. When visual hints are integrated into questions, the F1 score improves by at least 2.89\%, and the model achieves the best performance with 0.3442 by F1 and 0.2085 by BLEU when both ViLT and OFA hints are used. At final stage, adding image feature from ViT to question-hint sequences help improve the performance of previous models. Based on F1, these two combinations ConvS2S{\tiny~}+{\tiny~}ViLT{\tiny~}+{\tiny~}OFA and ConvS2S{\tiny~}+{\tiny~}ViT{\tiny~}+{\tiny~}ViLT{\tiny~}+{\tiny~}OFA are considered as our best methods on the public test set. 
Figure \ref{loss} depicts the gradual improvement in both training loss and testing loss as more image features and hints are added to the ConvS2S model. Knowledge-based ConvS2S (red line) does not catch the image context and thus have the highest loss. Though ConvS2S with ViT+VILT features does not obtained a competitive result on evaluation metrics, it gives the best loss among methods in the public test phase. In general, the optimal testing loss of methods is achieved between 14th and 20th epoch, then the models tend to be overfitting.



We manage to deploy two ensembles of ConvS2S using features from ViT combined with hints from {\tiny~}ViLT and {\tiny~}OFA, respectively, for the final evaluation on private test set. As shown in Table \ref{result_private}, the ConvS2S{\tiny~}+{\tiny~}ViT{\tiny~}+{\tiny~}OFA model obtained the better result, which is 0.4210 by F1 and 0.3482 by BLEU, and ranked $3^{rd}$ in the challenge. Table \ref{ranking} shows the final standing at the EVLSP-EVJVQA competition, in which our best model perform poorer 1.82\% and 1.39\% by F1 compared with the first and second place solutions. In terms of methodology, our approach comes in second place after the ViT{\tiny~}+{\tiny~}mT5 method, which has a large amount of pre-trained data. Overall, there is a gap between F1 and BLEU scores.

\begin{table}[H]
    \centering
    \small
    \setlength{\tabcolsep}{5pt}
    \renewcommand{\arraystretch}{1.2}
    \begin{tabular}{lcc}
    \toprule
        \textbf{Model} & \textbf{F1} & \textbf{BLEU}  \\ \midrule
        ConvS2S + ViT + ViLT &0.4053  &0.3228  \\
        \textbf{ConvS2S + ViT + OFA} & \textbf{0.4210}  & \textbf{0.3482}
  \\ \bottomrule
    \end{tabular}
    \caption{Performance on the private test set.}
    \label{result_private}
\end{table}

\begin{table}[!htbp]
\renewcommand{\arraystretch}{1.2}
\newcolumntype{L}{>{\centering\arraybackslash}m{7cm}}
\centering
\resizebox{\columnwidth}{!}{%
\begin{tabular}{clLccccc}
\toprule
\multirow{2}{*}{\textbf{No.}} & \multirow{2}{*}{\textbf{Team name}} & \multirow{2}{*}
{\textbf{Models}} & \multicolumn{2}{c}{\textbf{Public Test}} && \multicolumn{2}{c}{\textbf{Private Test}} \\\cmidrule{4-5} \cmidrule{7-8}
                             &                                     && \textbf{F1}         & \textbf{BLEU}      && \textbf{F1}         & \textbf{BLEU}       \\\midrule
1                            & CIST AI  &  ViT + mT5                          & 0.3491              & 0.2508             && 0.4392              & 0.4009              \\\midrule
2                            & OhYeah & ViT + mT5                             & 0.5755              & 0.4866             && 0.4349              & 0.3868              \\\midrule
\textbf{3}                            & \textbf{DS\_STBFL} & \textbf{ConvS2S+ViT+OFA}                  & \textbf{0.3390}     & \textbf{0.2156}    && \textbf{0.4210}     & \textbf{0.3482}     \\\midrule
4                            & FCoin & ViT + mBERT                               & 0.3355              & 0.2437             && 0.4103              & 0.3549              \\\midrule
5                            & VL-UIT & BEiT + CLIP + Detectron-2 + mBERT + BM25 + FastText                              & 0.3053              & 0.1878             && 0.3663              & 0.2743              \\\midrule
6                            & BDboi & ViT + BEiT + SwinTransformer
+ CLIP + OFA + BLIP                              & 0.3023              & 0.2183             && 0.3164              & 0.2649              \\\midrule
7                            & UIT\_squad & VinVL+mBERT                          & 0.3224              & 0.2238             && 0.3024              & 0.1667              \\\midrule
8                            & VC\_Internship & ResNet-152 + OFA                      & 0.3017              & 0.1639             && 0.3007              & 0.1337
\\\midrule9                            & Baseline & ViT + mBERT                      & 0.2924              & 0.2183             && 0.3346              & 0.2275\\\bottomrule       
\end{tabular}}
\caption{Our performance compared with other teams at VLSP2022-EVJVQA \cite{vlsp2022}}
\label{ranking}
\end{table}

\subsection{Performance Analysis}

According to the final result in the private test phase, the generated output from ConvS2S
+ViT+OFA model are chosen for further analysis. Generally, the model manages to generate answers with correct language with the input question.
\subsubsection{Quantitative analysis}
We randomly choose 100 samples from the generated result to perform quantitative analysis. The average length, vocabulary size, and the number of POS tags in the ground truth and generated answers are calculated for each language. Table \ref{quanti} shows the statistics of the ground truth answer compared with the predicted answer by the model.





\begin{table}[ht]
\centering
\begin{tabular}{llrr}
\toprule
Language&Stats.&Ground Truth&Predicted\\\midrule
\multirow{ 3}{*}{English} & Avg.length  & 3.74 & 6.18 \\
& Vocab. size & 78 & 72 \\
& \# POS tag  & 12 & 9 \\\midrule

\multirow{ 3}{*}{Vietnamese} & Avg.length  & 4.42 & 5.97 \\
& Vocab. size  & 97 & 101 \\
& \# POS tag &10  &9 \\\midrule

\multirow{ 3}{*}{Japanese} & Avg.length   & 4.67 & 8.43 \\
& Vocab. size & 77 & 83 \\
& \# POS tag  & 10 & 11 \\\midrule\midrule

\multirow{ 3}{*}{All} & Avg.length  &4.26 &6.78 \\
& Vocab. size &252 &256 \\
& \# POS tag  &14 &14 \\

\bottomrule
\end{tabular}
\caption{The quantitative statistic of 100 generated samples compared with the ground truth}
\label{quanti}
\end{table}

From Table \ref{quanti}, it can be seen that although the model gave the answers longer than the ground truth answers, the semantics is not as much as the ground truth. It can be seen from Table \ref{quanti} that the predicted answers in English have an average length higher than the ground truth answers. Also, the vocabulary in the generated answers is more than the original. In contrast, the number of POS tag components in the predicted answers is lower than the ground truth. This is similar to the answers in Vietnamese. For the Japanese, the characteristics of the predicted answers in average length and vocabulary size are the same as the two remaining languages. However, the number of POS tags in the predicted answers is more than in the ground truth answers. To make it clear, we propose three types of error on our model in Section \ref{quali_analysis}.

\begin{figure}[!ht]
    \centering
    \includegraphics[width=\textwidth]{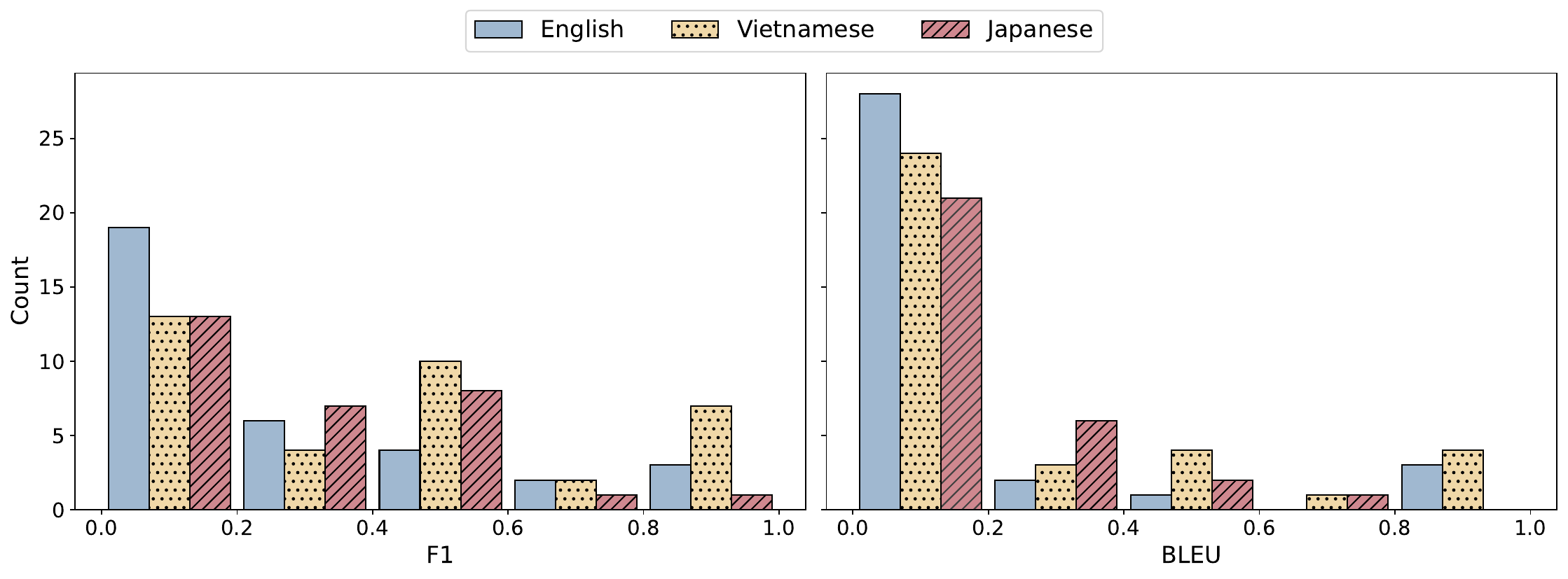}
    \caption{Distributions of F1 and BLEU scores for each language from 100 generated samples}
    \label{100score}
\end{figure}

In addition, Figure \ref{100score} illustrates the distributions of F1 and BLEU scores for each language. Generally, the histograms skewed to the right and the model  performs inconsistently across languages. The proportion of samples with F1 and BLEU scores less than 0.2 dominates the overall result across all three languages. In Vietnamese, the number of generated samples with F1 and BLEU scores greater than 0.4 is significantly higher than in other languages. Meanwhile, English and Japanese responses rarely score greater than 0.6 on both metrics, furthermore, no Japanese samples scoring greater than 0.8 in BLEU. This illustrates that our model faces numerous challenges in producing the desired responses, with specific limitations on each language.

\begin{figure}[!htbp]
\centering
\subfloat[]{%
  \includegraphics[width=0.8\textwidth]{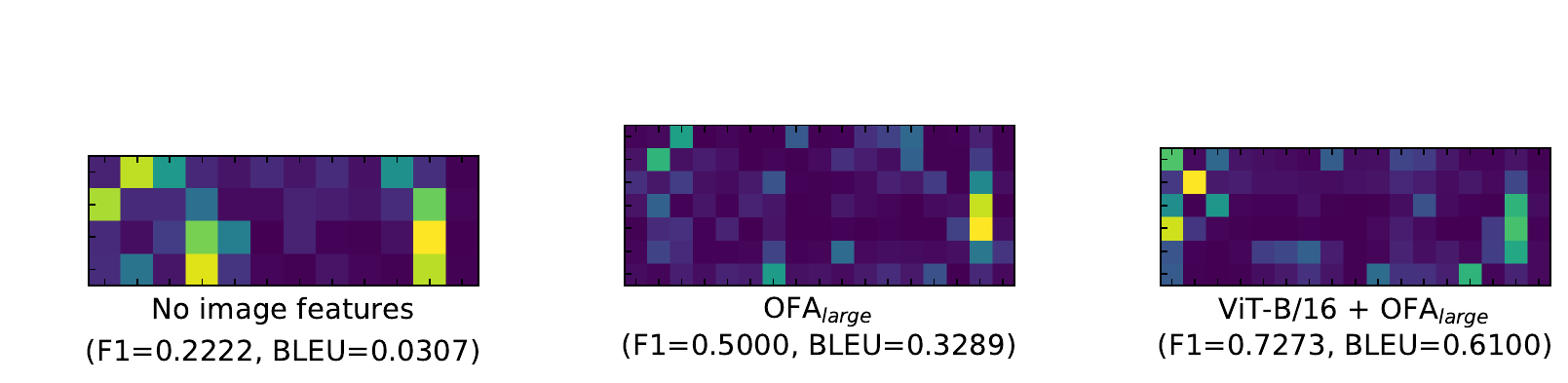}%
  \label{attn1}
}

\subfloat[]{%
  \includegraphics[width=0.8\textwidth]{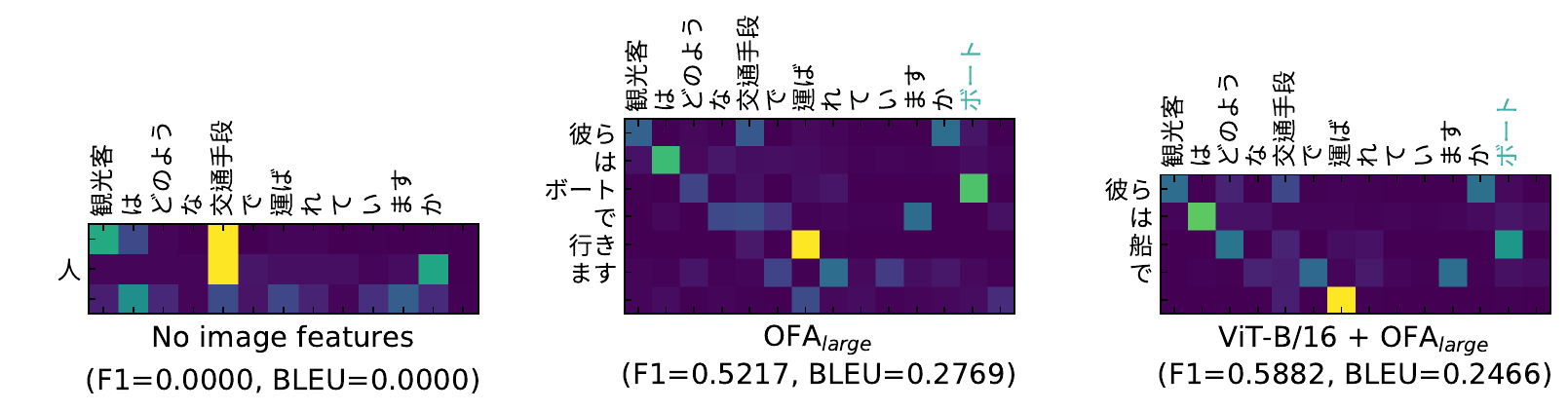}%
  \label{attn2}
}

\subfloat[]{%
  \includegraphics[width=0.8\textwidth]{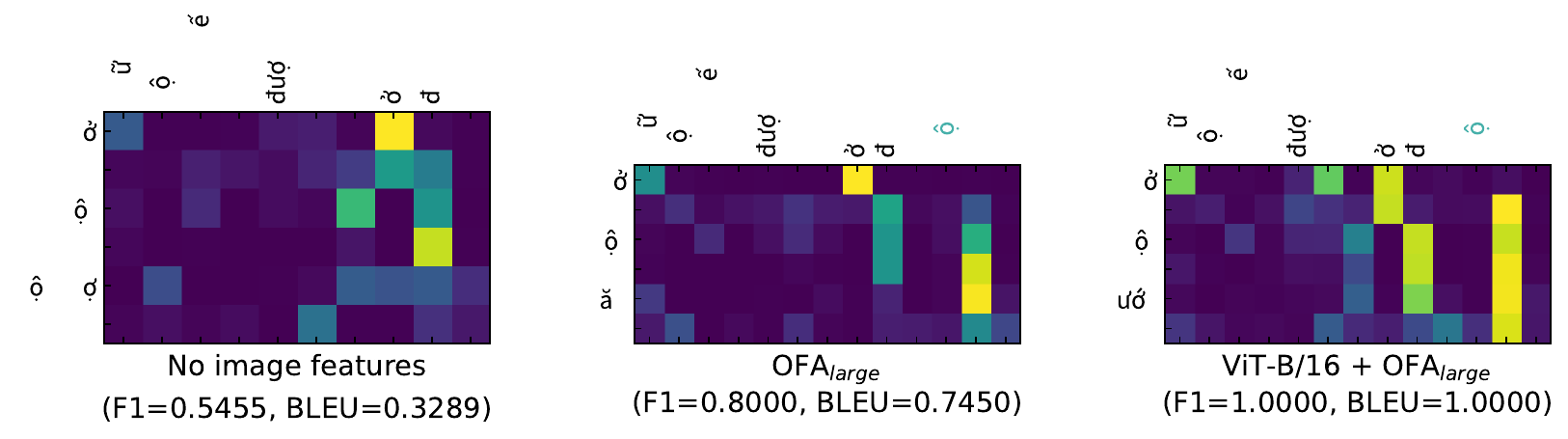}%
  \label{attn3}
}

\subfloat[]{%
  \includegraphics[width=0.8\textwidth]{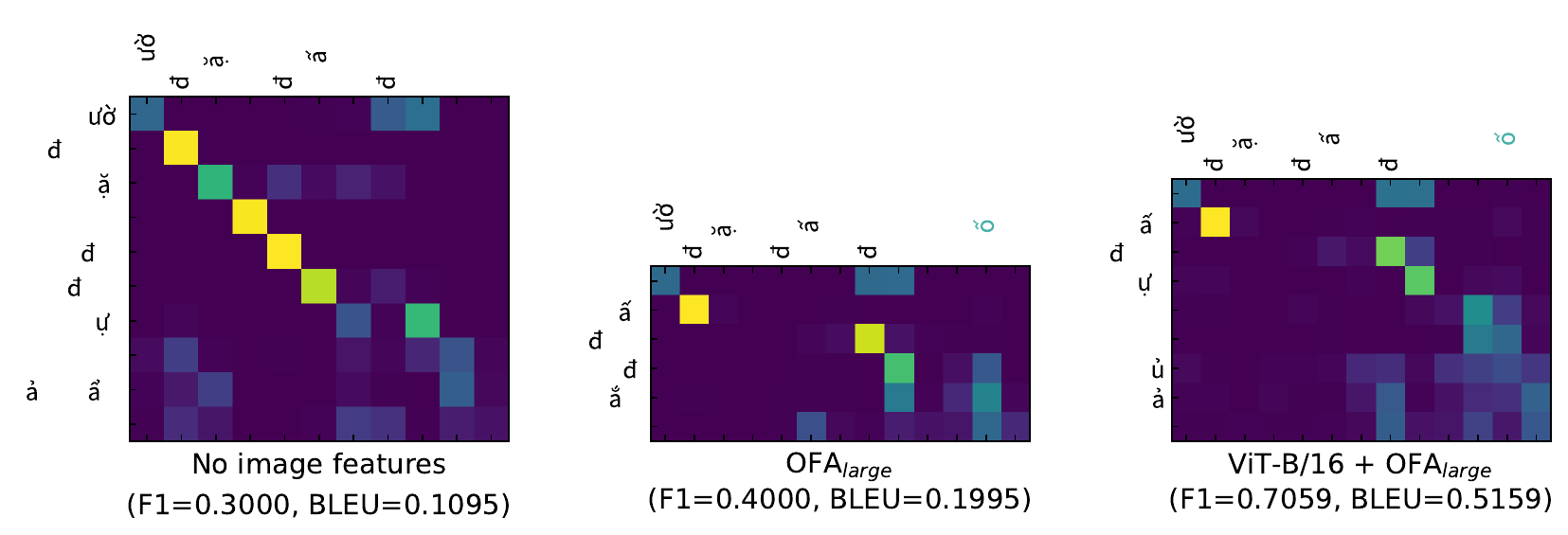}%
  \label{attn4}
}

\subfloat[]{%
  \includegraphics[width=0.8\textwidth]{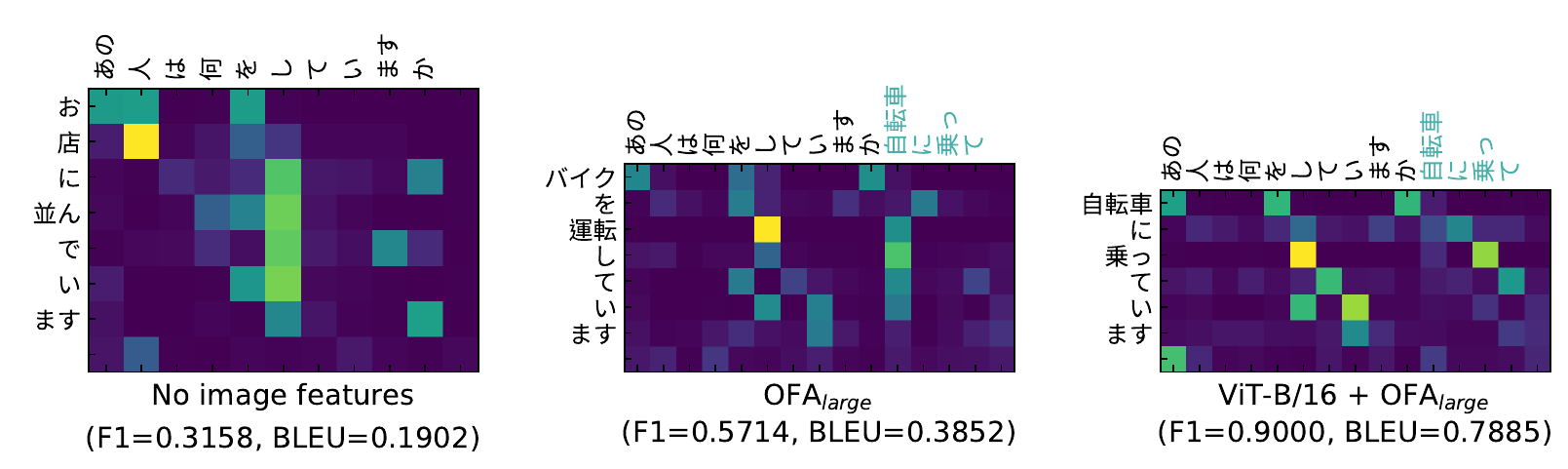}%
  \label{attn5}
}
\caption{Numerous samples of attention alignment from ConvS2S and the changes in attention when adding features from ViT and OFA models. The x-axis and y-axis of each plot correspond to the words in the question and the generated answer, respectively, while each pixel illustrates the weight $w_{ij}$ of the assignment of the j-th question word for the i-th
answer word.}
\label{attn}
\end{figure}

\subsubsection{Qualitative analysis}
\label{quali_analysis}
\paragraph{Attention visualization}

Figure \ref{attn} shows several samples of attention weights between each element from the generated answer with those in the input sequence that contains no image features, OFA hints, and ViT+OFA combined features, respectively. The visualization provided an intuitive way to discover which positions in the input sequence were considered more important when generating the target answer word. The brighter a pixel's color, the more important the word in the input sequence is in producing the respect answer word. The first heatmaps illustrate the case where no image information is used during training but only question. This is similar to Knowledged-based QA task where the model gives the answer solely based on the context of question. As a result, the generated answer is just simply a guess from ConvS2S model and has poor evaluation results on both metrics. Through attention visualization, we study that OFA hint is importance feature to model's attention as it provides the near-correct insight for the question and reduces the reliance on question words when generating the answer. This reduction in attention is not completely common for all question tokens, and it still depends on the importance of other elements in the whole sequence. However, in some cases, the model focuses too much on a specific element of the hint, which may lead to bias. ViT features has shown to control the affection of OFA hint, neutralizing it with other elements from question if hint appears to be off-topic. It may enhance the attention, making the model focus stronger on specific parts of the provided hint, for instance, the hint token ``nhà hàng'' (\textit{restaurant}) in Figure \ref{attn3} is given more attention when adding ViT image features. These features can also reduce the attention in one element and distributes concentration on other parts of the sequence. Figures \ref{attn1} and \ref{attn2} depict the reduction in hint attention into question elements, while Figures \ref{attn4} and \ref{attn5} show the changes in attention weight distribution among hint tokens.

\paragraph{Error analysis}
\begin{figure}[!ht]
\centering
\subfloat[Error Case I]{%
  \includegraphics[width=\textwidth]{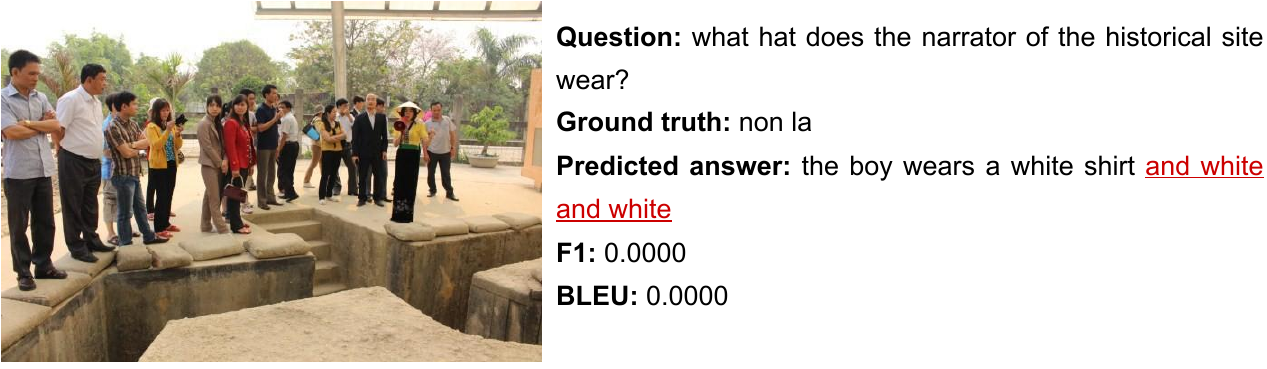}%
\label{fig:1a}}
\vspace{1em}
\subfloat[Error Case II]{%
  \includegraphics[width=\textwidth]{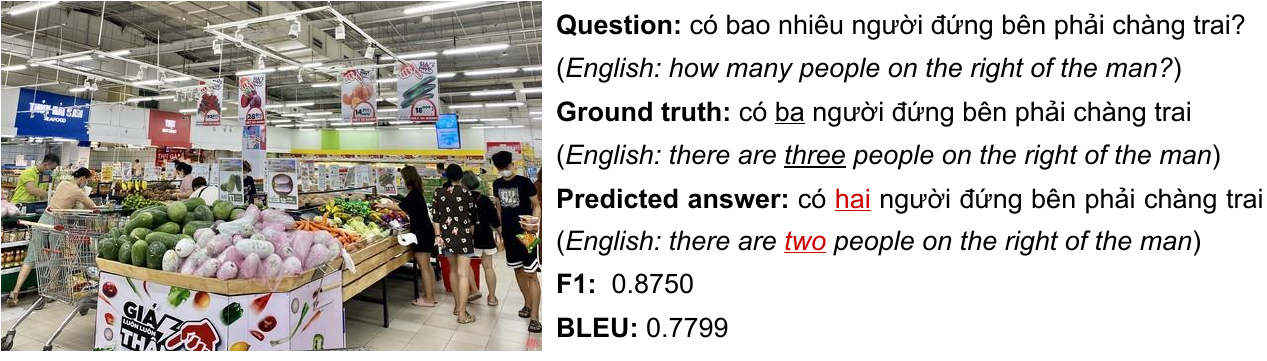}%
    \label{fig:1b}
}
\vspace{1em}
\subfloat[Error Case III]{%
  \includegraphics[width=\textwidth]{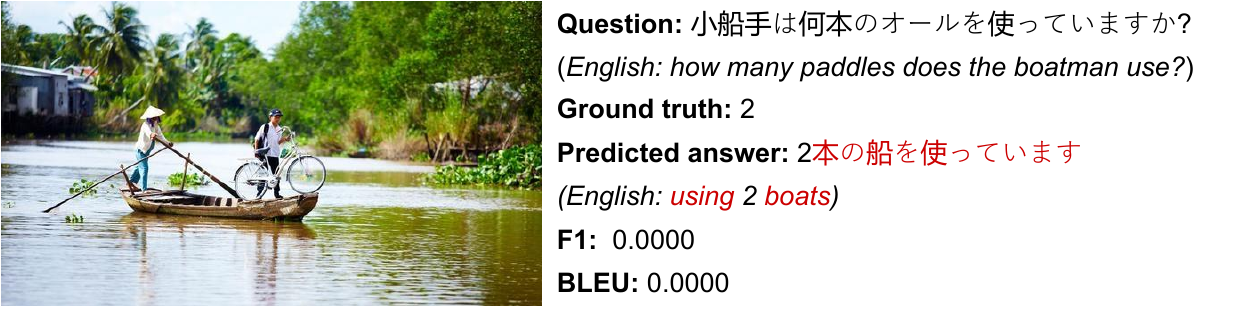}%
\label{fig:1c}}
\caption{Three typical error cases from generated results.}
\label{fig:1}
\end{figure}

For better understand the generation performance on the VQA task, we examine the generated answers of our best ensemble, ConvS2S
+ViT+OFA, to identify the limitations and analyze factors that may cause the model to perform poorly.
Through the error analysis process, various errors and mistakes have been pointed out in the outputs of the model. The typical examples of various types of errors are illustrated in Figure \ref{fig:1}. In summary, we divide these errors into three groups:

\begin{itemize}
    \item The generated answer does not match the question and has no correct tokens compared with the ground truth answer, as shown in Figure \ref{fig:1a}. This error case sometimes accompanied by text degeneration.
    \item The output response gives the wrong answer to the question but shares some insignificant tokens or has a similar structure with the ground truth answer, as shown in Figure \ref{fig:1b}, which significantly improves the evaluation score. This incorrect scenario exemplifies the limitation of the evaluation measures.
    \item The model managed to generate the correct key answer while also adding unnecessary information compared to the ground truth, which may lead to the response's meaning being distorted. 
    As shown in Figure \ref{fig:1c}, the model correctly predicted quantity but then added unnecessary tokens afterward, resulting in a low score on both evaluation metrics.
\end{itemize}


%% file: sections/6-conclusion.tex
\section{CONCLUSION}
\label{conclusion}
We have used the Convolutional Sequence-to-sequence network combined with the ViT and OFA model for our proposed system in the VLSP-EVJVQA task. The final results are 0.3390 on the public test set and 0.4210 on the private test set by the F1 score. From the result, we placed the $3^{rd}$ rank in the competition. Through errors analysis, various errors have been found in the output answer, which are our limitations in this study. In summary, there are factors that have significant impact on our solution for the multilingual VQA task: the diversity of each language, the translation performance, the effects of vision and language models and the generation capability of the core Seq2Seq model.

Our future research for this task is to improve the accuracy of the model in giving the correct answer by enriching the features from images and questions. Other SOTA vision-language and image models such as BEiT, DeiT and CLIP can be applied to assess the performance on UIT-EVJVQA dataset. Besides, from the proposed system, we will implement an intelligence chat-bot application for question-answering from images.